\documentclass[runningheads]{llncs}
\usepackage[T1]{fontenc}
\usepackage{enumitem}
\usepackage{graphicx}
\usepackage[hyphens]{url}
\usepackage{epigraph} 
\usepackage{amsfonts}
\usepackage[dvipsnames,svgnames,table]{xcolor}

\usepackage[colorlinks=true]{hyperref}
\hypersetup{
    linkcolor=blue,
    citecolor=blue,
    filecolor=blue,
    urlcolor=blue
}

\usepackage{url}
\begin{document}
\title{Beyond Detection: Rethinking Education in the Age of AI-writing}
\titlerunning{Beyond Detection: Rethinking Education in the Age of AI-writing}

% \author*[1,2]{\fnm{Maria} \sur{Marina}}\email{iauthor@gmail.com}

\author{Maria Marina\inst{1,2}%\orcidID{0009-0002-2274-4003}
\and Alexander Panchenko\inst{2,1}%\orcidID{0000-0001-6097-6118}
\and Vasily Konovalov\inst{1,2,3}%\orcidID{0000-0002-4745-4718}
}
% %
\authorrunning{M. Marina et al.}
\institute{AIRI \and Skoltech \and Moscow Institute of Physics and Technology\\
% \email{\{marina, konovalov\}@airi.net}}
\email{\{m.marina.scientia\}@gmail.com}}

\maketitle              % typeset the header of the contribution
\begin{abstract}
As generative AI tools like ChatGPT enter classrooms, workplaces and everyday thinking,  writing is at risk of  becoming a formality -- outsourced, automated and stripped of its cognitive value. But writing is not just output; it is how we learn to think. This paper explores what is lost when we let machines write for us, drawing on cognitive psychology, educational theory and real classroom practices. We argue that the process of writing -- messy, slow, often frustrating -- is where a human deep learning happens. The paper also explores the current possibilities of AI-text detection, how educators can adapt through smarter pedagogy rather than bans, and why the ability to recognize machine-generated language may become a critical literacy of the 21st century.  In a world where writing can be faked, learning can not.

% AI models become more sophisticated, their outputs increasingly mirror human language patterns, making detection more challenging. Theoretical analyses suggest that if AI-generated text distributions become indistinguishable from human text across their entire range, detection may become theoretically impossible. This challenges traditional education: How can we assess learning when AI can generate high-quality text? What should the educator’s role be in an AI-driven world? How do we ensure students develop critical thinking rather than relying passively on AI?

% This paper explores these challenges, focusing on two key areas: the (im)possibility of reliably detecting AI-generated text and the need to rethink both assessments and educators’ roles. To prepare students for an AI-integrated future, we must embed AI literacy into curricula, teaching both its capabilities and ethical implications. Traditional assessments must evolve, shifting from rote assignments to tasks that emphasize critical thinking, problem-solving, and knowledge application—skills AI struggles to replicate. Beyond this, educators must transition from information providers to facilitators of critical inquiry. Studies show that frequent AI users become better at detecting AI-generated content, even without training. This underscores the need for targeted instruction—both for educators, to enhance grading accuracy, and for students, to navigate misinformation effectively. By embracing these shifts, we can move beyond detection and create an education system that thrives in the age of AI-assisted writing.

\end{abstract}

\section{Introduction}

\epigraph{\itshape 
``I write because I don’t know what I think until I read what I say.''}{---Flannery O’Connor}

We are now seeing a growing trend where various roles are outsourcing their work to AI. Administrators are delegating condolence emails to automated tools, job applicants are using AI to craft tailored cover letters, marketers are generating entire ad campaigns with a few prompts, students are turning to AI to write their five-paragraph essays, and professors are relying on AI to grade those very same papers. ``When machines are grading machines, one should ask: what have we lost?''~\cite{jocelyn2024}.

Digital technologies, while offering unprecedented convenience, can also erode essential cognitive skills such as learning and memory retention~\cite{baron2021know}. When AI tools generate text on our behalf, they effectively short-circuit the deep thinking involved in the writing process. Writing is not just about producing words; it is a cognitive exercise that involves refining ideas, questioning initial assumptions, and often rewriting or restructuring entirely. This reflection fosters critical thinking and deeper understanding~\cite{baron2023}.

%This reflective process fosters critical thinking and strengthens long-term understanding.~\cite{baron2023}

When it comes to memory, recent research highlights a growing trend of relying on digital tools to store and retrieve information, an effect known as cognitive offloading. People are increasingly turning to the Internet as an external memory system, decreasing their reliance on internal recall~\cite{storm2017using}. This shift, while convenient, may come at the cost of weakening our natural memory functions.

Adding to this, individuals who searched online for information not only remembered less but also developed an inflated sense of their own knowledge~\cite{fisher2015searching}. In this study, participants who had looked up information on one topic (let us say topic X) were more likely to believe that they were knowledgeable about a completely unrelated topic (topic Y), simply because they felt confident that they could find the answer if needed. This illusion of understanding, fueled by the accessibility of online information, creates a false sense of expertise and undermines true learning. In contrast, while writing, the information is better memorized~\cite{silva2019writing}.

These findings align with broader concerns about how digital convenience shapes our cognitive habits. As psychologist Betsy Sparrow explains in her work on the ``Google Effect'', people are increasingly remembering how to find information rather than remembering the information itself~\cite{sparrow2011google}. This raises important questions about what happens to deep knowledge when we treat search engines as mental extensions rather than supplements.
\vspace{-1.5ex}
\section{Why do we still need to teach writing skills?}

In an era increasingly defined by automation and artificial intelligence, the role of writing in education is undergoing a profound shift. Rather than serving only as preparation for a career or a professional skillset, writing is being reframed -- particularly in some forward-thinking business schools -- as a foundational tool for living a thoughtful, intellectually engaged life~\cite{schatten2022will}. This liberal arts-inspired approach suggests that the true value of writing lies not in producing polished prose, but in what the act of writing enables: deliberate thinking, deeper reflection, and a clearer understanding of one’s own ideas.

This distinction is especially evident in the classroom, where students often fall into two broad camps. Some focus primarily on efficiency and outcomes -- grades, degrees -- and view writing as a hurdle to be cleared as quickly as possible. For them, generative AI tools are shortcuts that enable faster completion of assignments with minimal cognitive effort. But others see writing differently. These students treat it as a tool for discovery -- a means to figure out what they truly think and how best to express it. The process is not always easy. It involves confronting vague ideas, reshaping them through drafts, and wrestling with language to articulate precise, persuasive arguments. Yet, in doing so, they develop sharper, more nuanced reasoning skills and a deeper intellectual maturity~\cite{jocelyn2024}.

As Justin E.H. Smith points out, anti-intellectualism has long been a part of our cultural landscape, and the rise of AI only makes the struggle against it more difficult. A core purpose of education is to push back against this tendency -- to help students experience the value of learning from the inside rather than trying to convince them from the outside. Some students come eager to engage, but many are naturally resistant to learning efforts, especially when those efforts demand deep thinking. Still, as they work through their assignments, something often changes. They begin to sense the difference between uninformed opinions and carefully developed thoughts. Realization that clear, informed thinking feels different and has power is a pivotal moment in their intellectual growth. The concern with generative AI is that it can completely disrupt this process, offering a shortcut that skips over the essential struggle of learning how to think well~\cite{weinberg2025}.

Writing is the very act through which people learn to think clearly~\cite{pugh2014get}. Until one sits down to write, it is difficult to fully grasp the logic or coherence of one’s ideas. The process slows us down, forcing thoughts onto the page where they can be seen, questioned, and improved.  Scholars such as Hays~\cite{hays1983writer}, Lawrence~\cite{lawrence1972writing}, and Werse~\cite{werse2023will} reinforce this idea: writing is an iterative activity that deepens understanding by pushing us to organize and reorganize our thoughts for a particular audience. When that process is outsourced to AI, students may complete the task, but they forfeit the deeper cognitive and developmental benefits that come with struggling through it themselves. Naomi S. Baron states that relying on AI to generate content “diminishes opportunities to think out problems for ourselves”. She also mentioned the risk of a slippery slope, where students start letting generative AI control the content and style of their writing, leading to a diminished sense of ownership~\cite{wikipediaChatGPT2024}.

\vspace{-1.5ex}
\section{Impossibility of perfect detection of AI}

There are currently four main research directions aimed at addressing the challenge of AI-generated text detection. The first treats the task as a binary classification problem, using neural network-based detectors~\cite{openai2019gpt2,voznyuk-konovalov-2024-deeppavlov}. The second explores zero-shot detection methods~\cite{mitchell2023detectgpt}, relying on per-token log probabilities and thresholding to identify AI-generated text. The third line of work focuses on watermarking, which facilitates detection by embedding identifiable patterns into generated text. For example, a soft watermarking technique that categorizes tokens into ``green'' and ``red'' lists~\cite{kirchenbauer2023watermark}. A watermarked LLM is biased to select tokens from the green list, which is generated pseudo-randomly based on the prefix tokens. A detector can then classify a passage as AI-generated if it contains a disproportionately high number of green-list tokens.
Finally, information retrieval-based methods~\cite{krishna2023paraphrasing} detect AI-generated content by searching a database for semantically similar matches.

However, recent studies have shown that all of these approaches have significant limitations. \cite{sadasivan2023can} demonstrate that five rounds of recursive paraphrasing reduce detection accuracy on watermarked text to below 20\%. Similarly, the AUROC score of DetectGPT~\cite{mitchell2023detectgpt} drops from 96.5\% to 59.8\% after such attacks. Furthermore, the authors show that for any two distributions, where H is human text and M is AI-generated text — the best possible detector's AUROC score decreases as the total variation distance TV(M, H) between them shrinks. Another study finds that, with the exception of Pangram~\cite{pangram2024ai} (which is closed-source), all existing detection systems perform poorly under evasion strategies~\cite{russell2025people}.

As these findings show, there is currently no reliable, open-source solution for detecting AI-generated text. As human and AI-generated outputs grow more alike -- for instance, when TV < 0.2, -- even the best possible detectors yield unreliable performance (AUROC < 0.7)~\cite{sadasivan2023can}.

Moreover, even if watermarking were effective, it introduces several additional concerns. \textbf{Trade-offs in quality:} Watermarks may reduce fluency and quality, hurting user experience. Balancing imperceptibility and text quality remains a major challenge~\cite{nature2024editorial}. \textbf{Lack of standardization:} There is no widely accepted standard for watermarking AI-generated text. Different organizations developing their own methods could lead to compatibility issues and inconsistent detection outcomes across platforms~\cite{srinivasan2024}. \textbf{Ethical and privacy concerns.} Mandating watermarking could raise serious ethical and privacy issues, such as infringing on free expression or enabling surveillance. Additionally, watermarking may fail to address deeper problems like plagiarism or misinformation~\cite{lisinska2024}.
\vspace{-1.5ex}

\section{Use cases in secondary schools}

 As technology continues to evolve, educators find themselves navigating a rapidly changing landscape. One striking parallel is the introduction of generative AI tools like ChatGPT in writing classrooms. In the same way that calculators were once met with skepticism, but are now widely accepted as a necessary tool in math education, AI is becoming an integral part of teaching writing.
%Just as you wouldn’t ask a student to do complex arithmetic without a calculator, there’s no reason we shouldn’t allow AI to assist students in certain stages of the writing process. 
But like the gym, where results are gained not from watching others but from doing the work yourself, the real value lies in the student’s engagement with the process. It’s not about AI writing for students, but about using it to enhance their own thinking and writing skills.

%It's not about having AI write for them, but about learning to use AI as a tool that enhances their own thinking and writing skills.

All best practices used by both school teachers and university professors to teach writing  are divided into three categories.

% : “How To Use AI in Teaching Writing Effectively”, “Making It Harder for Students to Cheat Than Not Cheat” and “Other Ideas for Encouraging Authentic Writing”.

\paragraph{\bf How to use AI for teaching writing effectively.}
AI can serve as an assistant, not a replacement. It can help students in early stages, such as organizing their thoughts and generating ideas.
%, but it shouldn't be seen as a shortcut to bypass the hard work of writing.

\begin{itemize}[label=--]

\item \textit{AI-Assisted Outlining.} Cherie Shields, a high school teacher in Oregon, asked her students to use ChatGPT to generate outlines for essays comparing two 19th-century short stories. The students then wrote their essays by hand, deepening their understanding and refining their ideas. This strategy helped students not only interact with AI but also understand how to guide it to generate useful responses~\cite{roose2023don}.

\item \textit{Enhancing Visual Imagination.} Matt Cave, a head teacher at Willowdown, encourages students to feed their descriptive writing into AI to generate images. This helps them visualize what their words convey and compare it with their peers’ interpretations: ``they can then discuss with their classmates whether that was the image they expected to be in the reader’s head.''~\cite{adams2025}

\item \textit{Reflection and Revision.} One strategy involves asking students to submit two versions of their work: the original (non-AI), and the corrected (post-AI) version. In the post-AI version students have to comment on what mistakes AI has made and what the student is not sure about. This nice exercise teaches them to draw a line between legitimate and illegitimate use of resources~\cite{stackexchange2024}.
\end{itemize}

\paragraph{\bf Harder for the students to cheat than not to cheat.}
Instructors are developing clever ways to reduce the temptation to rely on generative AI -- not by banning it, but by designing assignments that are actually easier to complete honestly than to fake.

%By using ultra-specific prompts, staggered submission timelines, and creative constraints, educators are encouraging students to engage with their writing assignments in authentic, skill-building ways.
\vspace{-0.8ex}
\begin{itemize}[label=--]

\item \textit{Focused Mini-Assignments Over Full Papers.} Rather than assigning broad, full-length essays, some educators ask students to complete short, tightly focused writing tasks connected to key skills. For instance, students might be prompted to write a paragraph of lively prose, analyze a real-life observation, or turn a personal moment into a more universal insight. These concise tasks are harder to delegate to AI and often resonate more with students, making them more inclined to do the work themselves~\cite{bogost2024}.

\item \textit{Make Instructions Too Specific to Outsource.} Some instructors carefully craft prompts with detailed structure and formatting expectations that are tough to replicate with a simple AI query. For example, assignments may also be broken into weekly milestones to ensure consistency across drafts, making it much harder to outsource work submitted a week ago~\cite{stackexchange2024}.

%For example, students might be required to begin their paper with a paragraph that tackles two particular themes using a specific organizational style—like the "funnel method" discussed in class readings.

\item \textit{Constrained Vocabulary Challenge.} An effective vocabulary -- building activity involves giving students a list of 20 words and requiring them to use exactly 10 of them -- each only once in a short writing piece~\cite{stackexchange2024}.

%This forces students to make careful, deliberate choices about language and structure, creating a task that's difficult for AI to complete convincingly and rewarding for students to do on their own.

\end{itemize}

\paragraph{\bf Other ideas for encouraging authentic writing.}

Alongside structured prompts and clever assignment design, instructors are also experimenting with broader strategies to promote originality and student voice. 
%These methods— ranging from in-class freewriting to grading policies and current event-focused topics—help preserve student authenticity and minimize reliance on generative AI tools. 
%The goal isn’t just to prevent misuse, but to build confidence and critical thinking through real, personal engagement with the writing process.

\begin{itemize}[label=--]

\item \textit{In-Class Free-writing to Capture Voice.} Having students write short, unassisted paragraphs in class provides a snapshot of their natural writing voice. Over time, these samples help instructors track each student's style and recognize when later submissions deviate too far from it. This strategy not only discourages academic dishonesty but also gives students a chance to reflect on their developing voice~\cite{meyer2023generative}.

%As Nelson and Castello (2012) observed, voice emerges in the interaction between writer and reader, and instructors can use early writing exercises to support this growth.~\cite{nelson2012academic}

\item \textit{Personal Progress Grading.} Instead of comparing students to each other, some instructors evaluate work based on individual progress. By gathering an initial writing sample early in the term, teachers can measure each student’s growth over time. This approach eases anxiety, encourages effort, and allows students to take pride in their own development, and making shortcuts like AI-generated writing less appealing~\cite{stackexchange2024}.

\item \textit{Assign What AI Doesn’t Know.} One effective way to limit AI's usefulness in assignments is by asking students to write about recent news stories that fall outside AI training data. For example, students might be asked to analyze a current event that happened after a known AI model’s knowledge cutoff. Alternatively, they can exploit LLMs refusal to respond to politics-related queries. In one case, a student used AI to translate Burna Boy’s Nigerian Pidgin but it was necessary to rely on online forums for interpreting the political messages of the song~\cite{girdharry2023meaningful}.

%In one such case, a student noted that while AI could help translate Nigerian Pidgin of a song by Burna Boy, it was necessary to pointing to online forums where Nigerian fans offered their interpretations of the political messages of the song"~\cite{girdharry2023meaningful}

\end{itemize}

It is also important to note that banning access to generative technologies is counterproductive. Students who are determined to use these tools will likely find a way, but such restrictions tend to favor those whose first language is English, who can afford access to paid large language models or applications, and who are more digitally literate. Thus, increasing the gap between less and more priviliged students even more~\cite{furze2024}.
\vspace{-1.5ex}
\section{Recognizing AI-generated texts as a critical skill}
\vspace{-1.3ex}

Recent studies estimate that over 5\% of recently-published Wikipedia articles~\cite{brooks2024rise} and 10\% of PubMed abstracts published in 2024~\cite{kobak2024delving} were written by AI. This highlights the necessity to develop skills for detecting AI-generated content.

While the theoretical section demonstrates that evasion techniques can render AI-generated texts nearly undetectable, recent research~\cite{russell2025people} offers a more optimistic view for human detection. It shows that individuals who regularly use LLMs for writing tasks are significantly better at identifying AI-generated texts. In a study involving five such experienced annotators, only 1 out of 300 nonfiction articles was misclassified when decisions were made by majority vote.

Qualitative analysis revealed that these experts relied not only on lexical cues (such as ``AI vocabulary'') but also on more nuanced elements like clarity, originality, and formality -- features that are often difficult for automatic detectors to assess. This expertise proved robust across various models (including GPT-4o and Claude 3.5 Sonnet) and was unaffected by common evasion strategies, such as paraphrasing and humanization (i.e., prompts crafted using expert-derived instructions to simulate human writing). Remarkably, their detection capabilities extended even to content generated by the newly released o1-pro model.

If humans can be trained to recognize AI-generated texts, then this skill must become a core competency for both educators and students. A compelling real-world example was documented by Craig Meyer~\cite{meyer2023generative}:

%, who successfully identified an AI-written essay during the early days of ChatGPT’s popularity:

 \vspace{1.2ex}
 \textit{``The perfect sentences and perfect grammar. Finally, it dawned on me. Students don’t write like this; I don’t even write like this. It was too perfect. Yet there was something else, something much more important. It was dead. Lifeless. Perhaps even boring.
 Human writers tend to inject anecdotes, humor, slight digressions, stylistic preferences, and other perceptible creations of our genius. This paper had none of those, and once I saw it, I couldn’t unsee it. ''}
\vspace{-1.8ex}
\section{Conclusion}

The rise of generative AI confronts educators with a paradox: while these tools promise efficiency, they risk losing the very process that makes education transformative. When students bypass the struggle of writing, they forfeit the deep reflection that sharpens thinking, fosters memory, and stimulates intellectual independence. Instead of banning AI, we must redesign teaching to reward authentic voice, progress over perfection, and make cheating harder than learning. If we want students to write like humans, we must show them that writing \textit{makes} them human. And in a future where synthetic text is everywhere, the ability to think clearly, write originally, and detect what’s real won’t just be valuable -- it will be essential.

% P.S.: For writing of this article, AI assistant tools were only used for grammar checking and paraphrasing of several individual sentences.

\vspace{-2ex}

\bibliographystyle{splncs04}
\bibliography{short_bibliography}

\end{document}